\documentclass[letterpaper]{article} % DO NOT CHANGE THIS
\usepackage{aaai2026}  % DO NOT CHANGE THIS
\usepackage{times}  % DO NOT CHANGE THIS
\usepackage{helvet}  % DO NOT CHANGE THIS
\usepackage{courier}  % DO NOT CHANGE THIS
\usepackage[hyphens]{url}  % DO NOT CHANGE THIS
\usepackage{graphicx} % DO NOT CHANGE THIS
\usepackage{natbib}  % DO NOT CHANGE THIS AND DO NOT ADD ANY OPTIONS TO IT
\usepackage{caption} % DO NOT CHANGE THIS AND DO NOT ADD ANY OPTIONS TO IT
\usepackage{algorithm}
\usepackage{times}
\usepackage{helvet}
\usepackage{courier}
\usepackage{xcolor}
\usepackage{tabularx}
\usepackage{enumitem}
\usepackage{colortbl}
\usepackage{textcomp}
\usepackage{pifont}
\usepackage{multirow}
\usepackage{amssymb}
\usepackage{amsmath}
\usepackage{booktabs}
\usepackage{makecell}
\usepackage{booktabs}
\usepackage{siunitx}
\usepackage{pgfplots}
\usepackage{tikz}
\definecolor{textgray}{rgb}{0.7,0.7,0.7} 
\usepackage{algpseudocode}
\usepackage{float}
\usepackage{listings}
\usepackage{xcolor}
\usepackage[most]{tcolorbox}  % Required for creating the colored boxes
\usepackage{enumitem} % 引入新宏包，用于定制列表

\usepackage{newfloat}
\usepackage{listings}
\DeclareCaptionStyle{ruled}{labelfont=normalfont,labelsep=colon,strut=off} % DO NOT CHANGE THIS
\floatstyle{ruled}
\newfloat{listing}{tb}{lst}{}
\floatname{listing}{Listing}

\newtcolorbox{promptbox}[1]{ % <--- 这里加上 [1]
    enhanced,
    title=\textbf{#1}, % <--- 这里把具体的标题换成 #1
    colback=gray!5!white,
    colframe=gray!60!black,
    fonttitle=\bfseries,
    attach boxed title to top left={yshift=-2mm, xshift=3mm},
    boxed title style={
        colback=gray!60!black,
        arc=1.5mm,
    },
    arc=1.5mm,
    boxrule=1.5pt,
    breakable,
    width=0.49\textwidth, 
}

\newtcbtheorem[auto counter,
    list type=figure,
]{promptFig}{Figure}{
    enhanced,
    colback=gray!5!white,
    colframe=gray!60!black,
    fonttitle=\bfseries,
    attach boxed title to top left={yshift=-2mm, xshift=3mm},
    boxed title style={colback=gray!60!black, arc=1.5mm},
    arc=1.5mm,
    boxrule=1.5pt,
    breakable,
    label separator={:},
    width=0.49\textwidth,
}{fig} % 设置标签前缀，例如 fig:prompt4
\title{NOTAM-Evolve: A Knowledge-Guided Self-Evolving Optimization Framework with LLMs for NOTAM Interpretation}
\author{
    Maoqi Liu\textsuperscript{\rm 1},
    Quan Fang\textsuperscript{\rm 1}\thanks{Corresponding author},
    Yuhao Wu\textsuperscript{\rm 1},
    Can Zhao\textsuperscript{\rm 2},
    Yang Yang\textsuperscript{\rm 3,4},
    Kaiquan Cai\textsuperscript{\rm 3,4}
}
\affiliations{
    \textsuperscript{\rm 1}Beijing University of Posts and Telecommunications, Beijing 100876, China\\
    \textsuperscript{\rm 2}Aviation Data Communication Corporation, Beijing 100191, China\\
    \textsuperscript{\rm 3}Beihang University, School of Electronic and Information Engineering, Beijing 100191, China\\
    \textsuperscript{\rm 4}State Key Laboratory of CNS/ATM, Beijing 100191, China\\
    qfang@bupt.edu.cn
}

\begin{document}

\maketitle

\begin{abstract}
Accurate interpretation of Notices To Airmen (NOTAMs) is critical for aviation safety, yet their condensed and cryptic language poses significant challenges to both manual and automated processing. Existing automated systems are typically limited to ``Shallow Parsing,'' failing to extract the actionable intelligence needed for operational decisions. We formalize the complete interpretation task as ``Deep Parsing,'' a dual-reasoning challenge requiring both \textbf{dynamic knowledge grounding} (linking the NOTAM to evolving real-world aeronautical data) and \textbf{schema-based inference} (applying static domain rules to deduce operational status). To tackle this challenge, we propose \textbf{NOTAM-Evolve}, a self-evolving framework that enables a Large Language Model (LLM) to autonomously master complex NOTAM interpretation. Leveraging a knowledge graph-enhanced retrieval module for data grounding, the framework introduces a crucial closed-loop learning process where the LLM progressively improves from its own outputs, minimizing the need for extensive human-annotated reasoning traces. In conjunction with this framework, we introduce a new benchmark dataset of 10,000 expert-annotated NOTAMs. Our experiments demonstrate that NOTAM-Evolve achieves a 30.4\% absolute accuracy improvement over the base LLM, establishing a new state-of-the-art on the task of structured NOTAM interpretation. 
\end{abstract}

% Uncomment the following to link to your code, datasets, an extended version or similar.
% You must keep this block between (not within) the abstract and the main body of the paper.
\begin{links}
    \link{Code}{https://github.com/Estrellajer/NOTAM-Evolve}
\end{links}

\section{Introduction}
\label{sec:intro}

Notices To Airmen (NOTAMs) are official bulletins issued by aviation authorities to inform pilots and air traffic personnel of time-sensitive changes to airspace structure \cite{faa_notam_def}, airport facilities, or flight procedures. Unlike conventional technical documents, NOTAMs are written in highly condensed telegraphic language, often using specialized abbreviations, structured fields, and nonstandard syntax.

With over one million active NOTAMs issued annually worldwide \citep{morarasu2024aidriven}, accurate interpretation of these notices is essential for ensuring flight safety and operational efficiency. Misinterpretation can lead to missed warnings about closed runways, inoperative navigation aids, or restricted airspace—potentially resulting in costly delays or even safety incidents. However, manual NOTAM processing remains a labor-intensive and error-prone task.

\begin{figure*}[t!]
\centering
\includegraphics[width=0.99\textwidth]{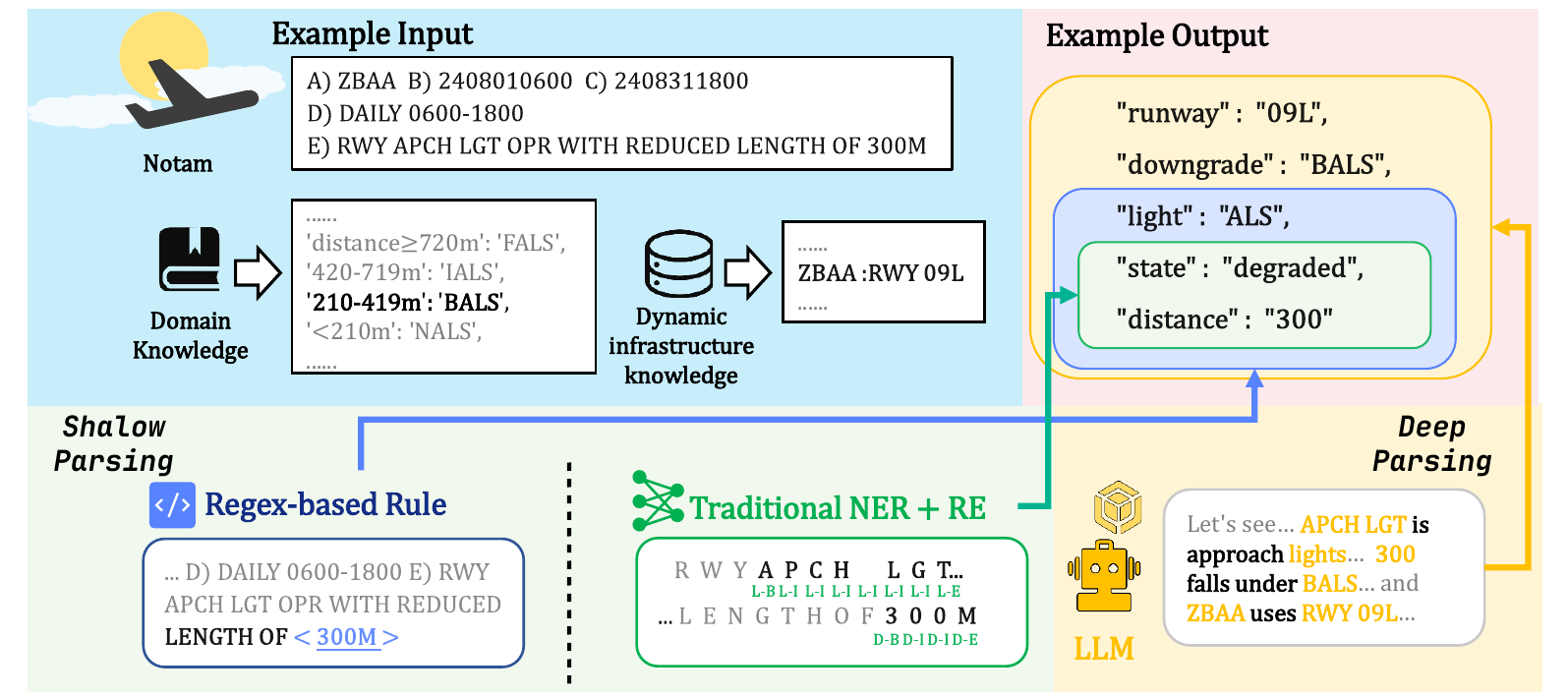}
\caption{An illustration of a NOTAM and its information parsing process. A NOTAM is a safety alert that reports flight hazards, and its original format ('Example Input') is unstructured text. Information parsing aims to convert this into structured data ('Example Output'). Past 'Shallow Parsing' approaches used techniques such as regular expressions or traditional NER. In contrast, 'Deep Parsing', as we define it, requires a model to combine domain knowledge with dynamic data for complex reasoning to understand the text's deep semantics.}
\label{fig:teaser}
\end{figure*}

Figure~\ref{fig:teaser} illustrates the core challenge of this interpretation task. Current automated systems, often relying on methods like rule-based pattern matching or traditional NER, are typically limited to solving what we term the task of "Shallow Parsing." These methods struggle to extract truly actionable intelligence, which still requires significant manual intervention. This necessitates a deeper level of interpretation, a capability we formalize as the task of "Deep Parsing." However, successfully performing this task is hindered by two fundamental difficulties that simplistic approaches cannot address.

The first challenge is \textbf{Dynamic Knowledge Grounding}. A NOTAM is not self-contained; as Figure~\ref{fig:teaser} illustrates, its interpretation requires grounding textual references in an external, dynamic knowledge base of aeronautical infrastructure. For instance, the airport code (\texttt{ZBAA}) serves as a key to retrieve that airport's specific configuration—such as its list of active runways (e.g., \texttt{RWY 09L})—from the knowledge base. This contextual data evolves over time and must be synchronized with the NOTAM's effective period~\citep{patel2023explainablepredictionqcodesnotams}.

The second challenge involves \textbf{Schema-Based Inference}. Deriving a NOTAM's true meaning requires schema-based reasoning beyond raw text extraction. Returning to the example in Figure~\ref{fig:teaser}, the statement \texttt{...REDUCED LENGTH OF 300M} provides a raw parameter, not an operational conclusion. An external schema, such as an ICAO rule set, is required to infer that this 300-meter system constitutes a "Basic Approach Lighting System (BALS)." This inferential step is crucial for translating raw facts into actionable intelligence and completing the deep parsing.

To address these dual challenges required for deep parsing, we propose NOTAM-Evolve, a self-evolving framework that enables LLMs to autonomously master complex NOTAM interpretation. The framework first employs a knowledge graph-enhanced retrieval module to ground interpretations in relevant aviation infrastructure data. Crucially, it then initiates a closed-loop learning process where the LLM's own outputs serve as training signals, enabling progressive improvement without requiring extensive human-annotated reasoning traces.

Our main contributions are threefold:
\begin{itemize}
    \item \textbf{Problem Formalization and Dataset}: We formally define the structured NOTAM interpretation task as a deep parsing challenge requiring both dynamic knowledge grounding and schema-based inference. We also introduce a comprehensive benchmark dataset of 10,000 globally sourced NOTAMs with expert annotations.
    
    \item \textbf{Self-Evolving Framework}: We propose a novel self-evolving framework that enables LLMs to autonomously master complex NOTAM interpretation through iterative preference optimization and consensus-based inference, without requiring extensive manual supervision.
    
    \item \textbf{Empirical Performance Leap}: Experimental validation demonstrates our optimized model achieves a 30.4\% accuracy improvement over base LLMs, establishing new state-of-the-art performance on this challenging aviation safety task.
\end{itemize}

\section{Related Work}
\label{sec:related_work}

\paragraph{NOTAM Parsing}
Automating the analysis of Notices to Airmen (NOTAMs) is a key application for NLP in aviation, aimed at reducing manual operational burdens \citep{mogillo-dettwilerfiltering, mi2022notam}. Initial research applied various techniques, from traditional NLP workflows like TF-IDF and NER for segmentation \citep{clarke2021natural} to transformer-based models for filtering and knowledge extraction from large, unlabeled corpora \citep{bravin2020automated, arnold2022knowledge}. Despite these pioneering efforts, persistent challenges such as ambiguous abbreviations, mismatches between semantics and operational practice, and regional variations remain significant hurdles to safety and efficiency \citep{morarasu2024aidriven}. Our work builds on these insights, proposing a more adaptive and robust LLM-based framework specifically designed to handle these complexities.Despite these pioneering efforts, persistent challenges such as ambiguous abbreviations, mismatches between semantics and operational practice, and regional variations remain significant hurdles to safety and efficiency \citep{morarasu2024aidriven}. Our work builds on these insights, proposing a more adaptive and robust LLM-based framework specifically designed to handle these complexities through knowledge grounding and iterative self-optimization.

\paragraph{Advances in Aviation NLP and LLMs}
More broadly, NLP is increasingly integral to enhancing flight operations and safety management. Recent applications range from improving flight trajectory prediction by integrating spoken instructions \citep{guo2024integrating}, graph-based modeling \citep{fan2024global}, and language modeling approaches \citep{luo2025large}, to developing agents for pilot training \citep{liu2024nlp} and automating the analysis of safety reports \citep{nanyonga2023aviation}. The progress in Large Language Models (LLMs) \citep{DBLP:journals/corr/abs-2303-18223} is central to these advancements. Modern transformer architectures \citep{brown2020languagemodelsfewshotlearners, Chowdhery2022PaLMSL}, especially when enhanced by parameter scaling \citep{Rae2021ScalingLM, scao2022bloom}, demonstrate strong few-shot learning capabilities well-suited to the sparsely labeled data common in aviation \citep{Xu2023FewshotMC}. Our work leverages several key LLM capabilities to tackle NOTAM parsing:

\textbf{Knowledge-Grounded Extraction and Reasoning.} Our framework's design is informed by LLM advancements in handling external knowledge and structured data. To enhance factual reasoning, we integrate a knowledge graph (KG), a strategy proven to improve reliability and reduce hallucinations in various domains \citep{ji2024retrieval, zhang2024making, chen2024sac, shi2024legal}. For the core task, we build upon modern information extraction—which utilizes in-context learning \citep{li2023codeie} and instruction tuning \citep{wang2023instructuie}—and tabular understanding, which has evolved from Text2SQL \citep{Zhong2017Seq2SQLGS} to methods like TableRAG \citep{Chen2024TableRAGMT}. Our work specifically targets the limitations of current innovations \citep{sainz2024gollie, li2024knowcoder} in handling the dynamic semantics and sparse schemas found in aviation.

\textbf{Optimization for Complex Instructions.} To manage the complex, constrained nature of NOTAMs, we employ curriculum learning. This approach is motivated by research showing that progressive learning \citep{mukherjee2023instruction, Luo2024HierarchicalNO} and specialized frameworks like Conifer \citep{Sun2024ConiferIC} are effective at improving a model's ability to follow complex instructions, which is central to our optimization strategy.

\section{Dataset Construction}
\label{sec:dataset}

\subsection{Dataset Overview}
We present a large-scale, comprehensive dataset of Notices To Airmen (NOTAMs) designed to support research on automated structured interpretation. This dataset contains 10,000 validated samples, averaging 39.2 words per notice and an average validity duration of 8.1 days.
The dataset covers a broad geographical distribution, with Asia as the top region (38.8\% of samples). The most frequent Q-Code is “Movement Area (M)” at 49.8\%. For evaluation, the dataset is divided into four subsets: Light (1,000 samples), Area (4,000 samples), Runway (2,500 samples), and Taxiway (2,500 samples).
The dataset was constructed through collection and aggregation of global NOTAM broadcasts, capturing real-world diversity and complexity. It has been curated with strict quality control to promote reliable and generalizable model development.
\subsection{Data Annotation}
Given the vast number of NOTAMs issued annually, we began by randomly sampling from the global NOTAM traffic for the entire year 2024. This approach ensures a manageable yet broadly representative volume, as substantiated by the statistics in Table~\ref{tab:categories}. The initial annotation schema was defined by expert aviation dispatchers, selecting foundational fields critical for operational efficiency and safety.
Our methodology employs a \textbf{non-extractive, inferential annotation scheme}. Unlike traditional information extraction datasets based on sequence labeling (e.g., BIO), annotators provided semantically correct values for each field regardless of whether these values were explicitly mentioned in the source text. For example, a NOTAM might imply a runway closure through technical jargon, and annotators explicitly assign the value "Closed" to the 
Runway Status
 field, as illustrated in Figure~\ref{fig:teaser}. This approach supports the development of models capable of genuine understanding and inference.
Annotation was independently performed by two expert dispatchers. To ensure consistency and quality, we computed Inter-Annotator Agreement (IAA), achieving a Krippendorff's Alpha of 0.96, indicating a very high level of reliability. All discrepancies were resolved by a third senior expert, thereby producing a gold-standard dataset.

\begin{table}[!htb]
\centering
\footnotesize \selectfont
\sisetup{round-mode=places, round-precision=1}
\begin{tabular}{lr}
    \toprule
    \textbf{Properties} & \textbf{Value} \\
    \midrule
    \multicolumn{2}{c}{\textbf{\textit{Overall Characteristics}}} \\
    \midrule
    \textbf{Total Samples} & 10,000 \\
    \textbf{Average Word Count} & \num{39.2} \\
    \textbf{Average Valid Days} & \num{8.1} \\
    \textbf{Top Region} & Asia (38.8\%) \\
    \textbf{Top Q-Code} & Movement Area (M, 49.8\%) \\
    \midrule
    \multicolumn{2}{c}{\textbf{\textit{Evaluation Task Distribution}}} \\
    \midrule
    \textbf{Light} & 1,000 \\
    \textbf{Area} & 4,000 \\
    \textbf{Runway} & 2,500 \\
    \textbf{Taxiway} & 2,500 \\
    \bottomrule
\end{tabular}
\caption{NOTAM dataset overview, including sample statistics and evaluation task distribution.}
\label{tab:categories}
\end{table}

\section{The NOTAM-Evolve Framework}
\label{sec:framework}
\begin{figure*}[ht!] % 使用 'ht!' 选项尝试将图片放置在当前位置
	\centering
    \includegraphics[width=16cm]{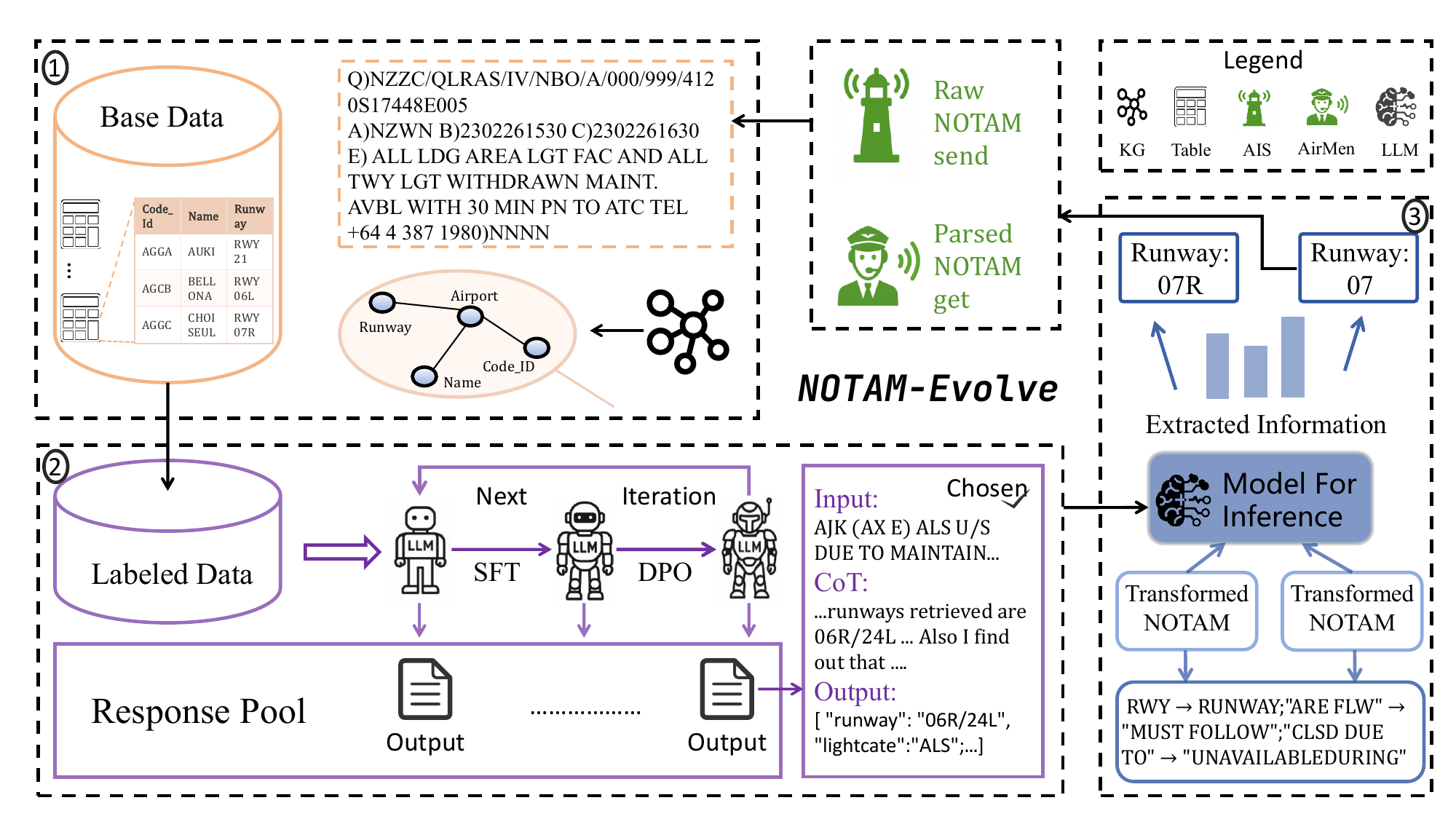} % 设置图片宽度以保持原始比例
	\caption{Overall framework of our proposed NOTAM-Evolve:
             (1) \textbf{Knowledge-Grounded Retrieval}: The final outputs are grounded in a set of base tables that represent real-world conditions, e.g., the number of runways at an airport. 
             (2) \textbf{Self-Optimizing Model Refinement}: Our foundational model gains proficiency in handling complex instructions within NOTAM analysis scenarios through iterative self-evolution combining supervised and preference optimization. 
             (3) \textbf{Multi-View Inference with Rewriting \& Voting}: We rephrase the original NOTAM without altering its core content and then extract information from multiple texts to determine the final answer via a voting mechanism.}
	\label{fig:framework} % 标签用于文内引用
\end{figure*}

\subsection{Problem Formulation}

Our primary objective is to extract structured aviation information from an input NOTAM text sequence $X = [x_1, \dots, x_n]$, by leveraging a collection of aviation reference tables $\mathcal{T} = \{T_1, \dots, T_m\}$ within a knowledge-enhanced generative framework. Formally, this task is defined as maximizing the conditional probability:
% The equation itself remains unchanged
\begin{equation}
p_\theta(Y \mid X, P, K) = \prod_{i=1}^m p_\theta(Y_i \mid X, P, K, Y_{<i}),
\label{eq:main_prob}
\end{equation}
% Minor rephrasing and reordering in the explanation of variables
where $Y = [Y_1, \dots, Y_m]$ denotes the target structured output sequence. The term $\theta$ represents the parameters of the LLM, $P$ encapsulates task-specific prompts and instructions, and $K = \kappa(X, \mathcal{T})$ corresponds to the factual knowledge retrieved from the aviation reference tables $\mathcal{T}$.

\subsection{Framework Overview}

As illustrated in Fig.~\ref{fig:framework}, our NOTAM-Evolve framework, also referred to as the Self-Evolving Framework, comprises three core stages: (1) The \textit{\textbf{Knowledge-Grounded Retrieval}} stage grounds predictions in aviation domain knowledge via dynamic table retrieval (TableRAG); (2) The \textit{\textbf{Self-Optimizing Model Refinement}} stage enables iterative self-improvement of the foundation model through a combination of supervised fine-tuning and adaptive preference learning; (3) The \textit{\textbf{Multi-View Inference}} stage ensures robust parsing via diversified input generation (rewriting) and consensus decoding (voting). This architecture allows NOTAM-Evolve to address key NOTAM analysis challenges, including knowledge grounding, error propagation, and stability.

\subsection{Knowledge-Grounded Retrieval}
The necessity for a retrieval-augmented approach stems from the dynamic nature of aviation data; core operational information, such as airport facility status and runway availability, is stored in tables that are periodically updated. A retrieval mechanism is therefore essential to ensure the Large Language Model (LLM) is grounded in the most current and factually accurate data. However, conventional retrieval methods are often insufficient, as they perform well only when table columns and their data possess clear, self-evident meanings. In the specialized aviation domain, this is rarely the case, as complex, implicit relationships are common. For instance, a simple query about a "runway closure" might fail to retrieve crucial, related information about dependent lighting systems or navigation aids because these connections are not explicitly defined in the table structure. To overcome this limitation, we introduce a domain-specific knowledge graph to provide the structured, real-world knowledge required for the LLM to better understand these complex relationships. Our proposed KG-TableRAG framework enhances the foundational TableRAG approach \citep{Chen2024TableRAGMT} by implementing a precise workflow: upon receiving a raw NOTAM, the LLM first generates a \texttt{Cypher} query to search the knowledge graph. The results from this graph query, which encapsulate relevant domain knowledge, are then concatenated with the original user query to form a new, enriched query. This new query is subsequently used to retrieve the most relevant information from the operational tables. Finally, this retrieved information is combined with the original NOTAM to create the definitive input for the language model's final processing, ensuring a factually consistent and context-aware output.

\subsection{Self-Optimizing Model Refinement}
\label{sec:sevo}

This stage iteratively refines the base model using a combination of supervised learning on correct predictions (self-supervised) and preference learning on error signals.

\noindent \textbf{Initialization Setup}
The process starts with:
\begin{itemize}[leftmargin=*,topsep=2pt,itemsep=2pt]
    \item \textbf{Data Partitioning}: An annotated dataset $\mathcal{D}_0 = \{(x \circ K, Y^*)\}$ is partitioned into training ($\mathcal{D}_{\text{train}}$) and test ($\mathcal{D}_{\text{test}}$) sets (e.g., 8:2 ratio). Here $x$ is the NOTAM text, $K = \kappa(x, \mathcal{T})$ is retrieved knowledge, and $Y^*$ is the ground truth structured output.
    \item \textbf{Base Model}: An initial model $\pi_0$, typically an untuned open-source LLM ($\pi_{\text{base}}$).
    \item \textbf{Response Pool}: An indexed set $\mathcal{R}$, initially empty, to store input-output pairs $(x, Y^*, \hat{Y})$ generated across iterations $e$.
\end{itemize}

\noindent \textbf{Iterative Optimization Loop}
Each iteration $e$ (from $1$ to a maximum $E$) involves intertwined SFT and DPO stages (See Fig.~\ref{fig:framework} and Algorithm in the supplementary material.

First, using the current model $\pi_e$, responses $\hat{Y}^{(e)}$ are generated for inputs $x \in \mathcal{D}_{\text{train}}$. These responses are compared with $Y^*$ to label them as correct or incorrect, and the repository $\mathcal{R}$ is updated. The error rate for an input $x$, $\xi(x)$, is estimated as the fraction of the last $K'$ generated responses that were incorrect:
\begin{equation}
\xi(x) = \frac{\sum_{k=1}^{K'} \mathbb{I}(\hat{Y}^{(k)} \neq Y^*)}{K'}
\label{eq:error_rate_def}
\end{equation}
where $K'$ is a hyperparameter defining the lookback window.

Next, supervised fine-tuning (SFT) is performed. Correct input-output pairs $(x \circ K, Y^*)$ are extracted from $\mathcal{R}$ to form the SFT dataset $\mathcal{D}_{\text{SFT}}^{(e)}$. The model $\pi_e$ is fine-tuned by minimizing the standard negative log-likelihood loss (where $m$ is the target sequence length):

\begin{equation}
\mathcal{L}_{\text{SFT}}^{(e)} = 
-\,\mathbb{E}_{(x,Y^*) \sim \mathcal{D}_{\text{SFT}}^{(e)}} 
\Big[ \sum_{i=1}^m \log \pi_\theta(Y_i^* \mid x\!\circ\! K,\, Y_{<i}^*) \Big]
\label{eq:sft_loss}
\end{equation}

Following SFT, the dynamic preference optimization (DPO) stage begins. A preference dataset $\mathcal{D}_{\text{pref}}^{(e)}$ is constructed by sampling triples $(x, y^*, y^-)$ from $\mathcal{R}$, where $y^*$ is a known correct response and $y^-$ is incorrect for input $x$. Dynamic data augmentation is applied for inputs $x$ with a high error rate ($\xi(x) \ge \tau$, where $\tau$ is a threshold), generating $N_{aug}$ semantic-preserving variants $\mathcal{V}_x$. Corresponding preference triples $(v, y^*, y^-)$ for $v \in \mathcal{V}_x$ are added to form the full DPO dataset:
\begin{equation}
\mathcal{D}_{\text{DPO}}^{(e)} = \mathcal{D}_{\text{pref}}^{(e)} \cup \bigcup_{\substack{x \in \mathcal{D}_{\text{train}} \\ \xi(x) \geq \tau}} \left\{(v, y^*, y^-) \mid v \in \mathcal{V}_x \right\}
\label{eq:augment}
\end{equation}
Weighted curriculum learning is implemented when sampling from $\mathcal{D}_{\text{DPO}}^{(e)}$. The sampling weight $w_e(x)$ for input $x$ at iteration $e$ adaptively focuses on harder examples using the error rate $\xi(x)$ and a curriculum schedule $\alpha_e = \min(e/E, 1)$. Let $N=|\mathcal{D}_{\text{DPO}}^{(e)}|$, $\beta_{\text{weight}}$ controls error emphasis, and $E$ is the total scheduled iterations:

\begin{equation}
w_e(x) = (1-\alpha_e)\tfrac{1}{N}
+ \alpha_e\, \frac{\exp(\beta_{\text{weight}}\xi(x))}{
\sum_{j=1}^N \exp(\beta_{\text{weight}}\xi(x_j))}
\label{eq:curriculum}
\end{equation}

This transitions sampling from uniform towards error-weighted as iterations progress.
Finally, the DPO loss is optimized using the SFT-updated model as the policy $\pi_\theta$ and the model from the start of the iteration $\pi_e$ as the reference $\pi_{\text{ref}}$. Samples $(x, y^*, y^-)$ are drawn according to $P_e(x) \propto w_e(x)$:
% --- Using multline for the long DPO loss formula ---
\begin{multline}
\mathcal{L}_{\text{DPO}}^{(e)} = -\mathbb{E}_{\substack{(x,y^*,y^-) \\ \sim P_e(x)}} \bigg[ \log \sigma \bigg( \\
 \beta_{\text{DPO}} \log \frac{\pi_\theta(y^*|x)}{\pi_{\text{ref}}(y^*|x)}
 - \beta_{\text{DPO}} \log \frac{\pi_\theta(y^-|x)}{\pi_{\text{ref}}(y^-|x)} \bigg) \bigg]
\label{eq:dpo_loss}
\end{multline}
Here, $\beta_{\text{DPO}}$ is the DPO hyperparameter. The sigmoid function $\sigma(\cdot)$ transforms the scaled log-probability difference into a probability [0, 1], representing the preference likelihood ($y^*$ over $y^-$), enabling direct learning from preferences. The model after DPO becomes $\pi_{e+1}$.

The iterative loop terminates when the model $\pi_{e+1}$ achieves a target accuracy $\eta$ on the test set $\mathcal{D}_{\text{test}}$:
\begin{equation}
\frac{1}{|\mathcal{D}_{\text{test}}|} \sum_{(x,Y^*) \in \mathcal{D}_{\text{test}}} \mathbb{I}(\pi_{e+1}(x \circ K) = Y^*) \geq \eta
\label{eq:termination}
\end{equation}
where $\pi_{e+1}(x \circ K)$ is the model's prediction.

Empirical results show that the framework achieves commercial \textsc{SOTA}-level NOTAM parsing accuracy within 3-5 iterations without model distillation.

\begin{table*}[h]
  \centering
  \begin{tabular}{>{\bfseries}lccccc}
    \toprule
    \textbf{Model} & \textbf{Light} & \textbf{Area} & \textbf{Runway} & \textbf{Taxiway} & \textbf{AVG} \\
    \midrule
    \multicolumn{6}{l}{\textbf{Popular Models}} \\
    \midrule
    Regex Template Rule-based Matching & 0.370 & 0.491 & 0.443 & 0.396 & 0.425 \\
    UIE \citep{Lu2022UnifiedSG} & 0.270 & 0.380 & 0.320 & 0.430 & 0.350 \\
    qwen2.5-7B \citep{Yang2024Qwen25TR} & 0.560 & \underline{0.777} & 0.412 & 0.748 & 0.624 \\ 
    Mistral-7B \citep{Jiang2023Mistral7} & 0.405 & 0.655 & 0.588 & 0.492 & 0.535 \\ 
    Llama3.1-8B-instruct \citep{Dubey2024TheL3} & 0.440 & 0.476 & 0.392 & 0.490 & 0.450 \\ 
    Deepseek-R1-Distill-Qwen-7B & 0.410 & 0.484 & 0.446 & 0.492 & 0.458 \\
    qwen2.5-7b-instruct (SFT) & 0.590 & 0.793 & 0.730 & 0.864 & 0.744 \\ 
    Deepseek-R1-Distill-Qwen-7B (SFT) & 0.18 & 0.226 & 0.236 & 0.204 & 0.212 \\ 
    \textbf{NOTAM-Evolve (ours)} & \underline{0.620} & 0.725 & \underline{\textbf{0.836}} & \underline{0.868} & \underline{0.762} \\ 
    \midrule
    \multicolumn{6}{l}{\color{textgray}\textbf{Commercial Models}} \\ % 分组标题变灰
    \midrule
    \color{textgray}GPT-4o \citep{Achiam2023GPT4TR} & \color{textgray}0.605 & \color{textgray}0.851 & \color{textgray}0.770 & \color{textgray}0.914 & \color{textgray}0.785 \\ 
    \color{textgray}Deepseek-R1 \citep{DeepSeekAI2025DeepSeekR1IR} & \color{textgray}\underline{\textbf{0.725}} & \color{textgray}\underline{\textbf{0.871}} & \color{textgray}\underline{0.792} & \color{textgray}\underline{\textbf{0.924}} & \color{textgray}\underline{\textbf{0.828}} \\
    \bottomrule
  \end{tabular}
  \caption{\label{performance-table}
    Performance comparison on four NOTAM analysis tasks. Models are grouped into Popular (including traditional methods and open-source LLMs) and Commercial (references). \underline{Underlined}: Best result within the Popular Models group or the Commercial Models group, respectively. \textbf{Bold}: Overall best result across all models.
  }
\end{table*}

\subsection{Multi-View Inference with Rewriting \& Voting}

Standard parsing paradigms struggle with NOTAM analysis due to models' limited complex instruction-following, often causing structural output errors. Particularly for edge cases where minor reasoning path variations could determine correctness, we observe that the baseline model ($\pi_{\text{R1}}$) generates inconsistent predictions despite demonstrating partial comprehension. To mitigate instability and preserve domain integrity, we use input diversification with consensus-based decoding (rewriting and voting). The approach begins with generating $N=5$ semantically-equivalent NOTAM variants through controlled paraphrasing that strictly maintains original aviation terminology (e.g., preserving "RWY" abbreviations), spatiotemporal constraints, and safety-critical numerical values. Each variant undergoes independent model processing to yield candidate structured outputs $\{\hat{Y}^{(k)}\}_{k=1}^N$, followed by majority voting to determine the final prediction $\hat{Y}_{\text{final}} = \arg\max_{Y} \sum_{k=1}^N \mathbb{I}(Y = \hat{Y}^{(k)})$. The paraphrasing mechanism combines lexical substitution (e.g., "\texttt{CTAM}" $\leftrightarrow$ "Controller Advisory Message"), syntactic restructuring through voice alternation, and contextual expansion with optional ICAO phraseology clarifications. Experimental validation in Section~\ref{sec:results} demonstrates this technique's effectiveness, achieving a 5\% accuracy improvement.

\section{Experiments}
\label{sec:experiments}

\subsection{Experimental Setup}

\noindent\textbf{Datasets \& Evaluation Protocol.}
For our evaluation, we leverage the specialized NOTAM dataset detailed in Section~\ref{sec:dataset}. A prediction is considered correct only if it exactly matches the ground truth in both format and all annotated field values.

\noindent\textbf{Baselines.} We benchmark our framework against several categories of baselines, as detailed in Table~\ref{performance-table}. These include: 1) traditional methods (a Regex-based system and the UIE information extractor \citep{Lu2022UnifiedSG}); 2) open-source LLMs (Qwen2.5-7B \citep{Yang2024Qwen25TR}, Mistral-7B \citep{Jiang2023Mistral7}, Llama3.1-8B-Instruct \citep{Dubey2024TheL3}, and our base model, DeepSeek-R1-Distill-Qwen-7B); 3) their SFT counterparts where applicable (Qwen2.5-7B-Instruct and DeepSeek-R1-Distill-Qwen-7B (SFT)); and 4) high-performance commercial models as reference points (GPT-4o \citep{Achiam2023GPT4TR} and DeepSeek-R1 \citep{DeepSeekAI2025DeepSeekR1IR}). To ensure a fair comparison, all LLMs were evaluated using identical inputs, which consist of the same domain-specific prompts and information retrieved by our Knowledge-Grounded Retrieval module.

\noindent\textbf{Implementation Details.} Our implementation is based on the DeepSeek-R1-Distill-Qwen-7B model. Fine-tuning (both standard SFT and our iterative optimization) was performed using the Unsloth framework with its recommended configurations. Further details on the knowledge graph and the prompts used can be found in the supplementary material. All experiments were conducted on a single NVIDIA A800-80GB-PCIe GPU.

\subsection{Main Results}

We evaluated our optimized model, \textit{DeepSeek-R1-Distill-Qwen-7B (ours)}, against the baselines on four key NOTAM analysis tasks. Table~\ref{performance-table} summarizes these results. Our model achieves a high average score (AVG), significantly outperforming traditional methods. Crucially, it obtains a 30.4\% absolute improvement over its base model (DeepSeek-R1-Distill-Qwen-7B), directly validating the effectiveness of our optimization pipeline. Furthermore, our model surpasses other tested open-source LLMs (e.g., Mistral-7B, Llama3.1-8B-Instruct) and the best-performing SFT baseline (Qwen2.5-7B-Instruct).

Notably, employing SFT alone can degrade a model's reasoning capabilities. This aligns with findings that fine-tuning on datasets lacking Chain-of-Thought (CoT) rationales can impair reasoning performance \citep{lobo2024impact}. Finally, as shown in Table~\ref{performance-table}, our model, despite its significantly smaller parameter count, achieves performance comparable to that of GPT-4o and DeepSeek-R1. This result is not merely a matter of efficiency; it is a critical step toward practical deployment, as the operational, security, and cost constraints of aviation often preclude the use of closed-source, third-party APIs.

\subsection{Ablation Study}
\label{sec:results}

We conduct systematic ablation analyses to validate our design choices by removing key components: (1) KG-TableRAG knowledge integration and (2) the Multi-View Inference mechanism. 

Table~\ref{tab:final_centered} demonstrates that removing KG-TableRAG (-KG) causes a 2.2\% performance drop (0.740 vs 0.762 AVG), particularly affecting knowledge-dependent tasks like Q-code mapping. Removing Multi-View Inference (-Multi-View) results in a larger 4.1\% decline (0.721 AVG), confirming its critical role in handling prediction instability and edge cases. When both components are removed, performance drops to the lowest level (0.690 AVG), demonstrating their complementary necessity. 

\begin{table}[htbp]
  \centering
  \setlength{\tabcolsep}{0.8em}
  \begin{tabular}{@{}ccS[table-format=1.3]@{}}
    \toprule
    \textbf{KG-TableRAG} & \textbf{Multi-View} & \textbf{AVG} \\
    \midrule
    $\checkmark$ & $\checkmark$ & 0.762 \\
    $\checkmark$ & $\times$     & 0.721 \\
    $\times$     & $\checkmark$ & 0.740 \\
    $\times$     & $\times$     & 0.690 \\
    \bottomrule
  \end{tabular}
  \caption{Ablation Study Results with KG-TableRAG and Multi-View Inference Components.}
  \label{tab:final_centered}
\end{table}

\begin{figure}[!htbp]
\centering
\begin{tikzpicture}
\begin{axis}[
    ybar,
    bar width=11pt,
    enlargelimits=0.15,
    legend style={at={(0.5,-0.5)},
                  anchor=north,legend columns=-1},
    symbolic x coords={Light, Area, Runway, Taxiway},
    xtick=data,
    nodes near coords,
    nodes near coords align={vertical},
    every node near coord/.append style={
        font=\tiny,
        anchor=south,
    },
    width=\columnwidth,
    height=0.45\columnwidth,
    title={Performance comparison across iterations},
    ymin=40,
    ymax=100,
    enlarge x limits={abs=0.8cm},
]

% Iteration 1 data
\addplot coordinates {(Light, 45) (Area, 63) (Runway, 78.8) (Taxiway, 64.6)};
% Iteration 2 data
\addplot coordinates {(Light, 54) (Area, 72) (Runway, 84.2) (Taxiway, 81.2)};
% Iteration 3 data
\addplot coordinates {(Light, 62) (Area, 73) (Runway, 83.6) (Taxiway, 86.8)};

\legend{Iteration 1, Iteration 2, Iteration 3}
\end{axis}
\end{tikzpicture}
\caption{Iterative Optimization Performance (Accuracy \%) across NOTAM Categories.}
\label{fig:performance-comparison-iterations}
\end{figure}
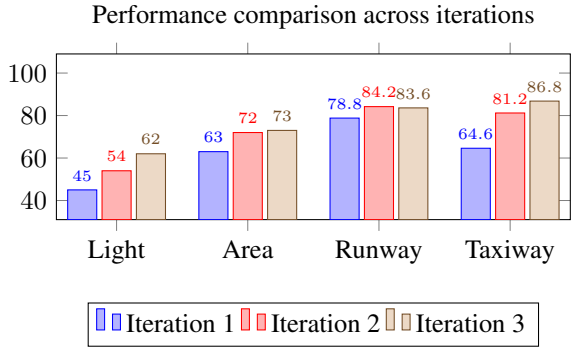

Figure~\ref{fig:performance-comparison-iterations} validates our iterative Self-Optimizing Model Refinement strategy, showing consistent improvements across all categories over three iterations. Notably, complex tasks like Taxiway accuracy improved from 64.6\% to 86.8\%, while Light accuracy increased from 45\% to 62\%. Collectively, these results confirm that each design choice is not only beneficial on its own, but that their iterative application is key to maximizing performance.

\subsection{Complexity Analysis}

We analyze the computational characteristics of our iterative optimization framework. The preference pair creation follows quadratic scaling modulated by accuracy progression:

\begin{equation}
|\mathcal{D}_{\text{pref}}^{(t)}| \approx 9K^2t^2(1-\eta)
\end{equation}

where $\eta$ denotes global accuracy (45\% → 62\% over 3 iterations), causing the error suppression term $(1-\eta)$ to decrease from 0.55 to 0.38.

\begin{table}[htbp]
\centering
\begin{tabular}{l@{\hspace{1.2em}}ccc}
\toprule
Metric & Iter.1 & Iter.2 & Iter.3 \\
\midrule
Theoretical pairs & 2,415 & 5,915 & 11,320 \\
Effective pairs & 1,449 & 3,549 & 6,792 \\
Time (h) & 0.58 & 1.5 & 3.2 \\
Scale factor & 1.0× & 2.6× & 2.1× \\
\bottomrule
\end{tabular}
\caption{Iterative Complexity Metrics with Scaling Factors}
\label{tab:complexity_metrics}
\end{table}

The computational cost per iteration is governed by:
\begin{equation}
\mathcal{T}_{\text{DPO}}^{(t)} = E \cdot |\mathcal{D}_{\text{pref}}^{(t)}| \cdot \mathbb{E}_{w_e(x)}[1/P_e(x)]
\end{equation}

where curriculum sampling weights $w_e(x)$ prioritize harder examples with higher error rates.

Three mechanisms suppress theoretical $O(t^2)$ scaling to observed 2.3× average growth: 1) Error threshold filtering removes 40\% of low-difficulty samples, 2) Curriculum sampling reduces effective batch size by 38\%, and 3) Accuracy saturation limits error generation through $(1-\eta)$ decay.

The framework maintains practical tractability with convergence achieved in 3 iterations at 62\% accuracy. Total wall-clock time ranges from 35 minutes to 3.2 hours on NVIDIA A800 GPUs.

\subsection{Case Study}
% In your Results section...

This case study illustrates our framework's advantage in reasoning about implicit, hierarchical relationships in NOTAMs, where high-level restrictions affect unmentioned components.

Consider a NOTAM for airport AGGC:
\begin{quote}\small
\texttt{E) CHOISEUL L BAY AIRPORT CLOSED TO ALL OPERATIONS...}
\end{quote}
Correctly interpreting this airport-wide closure means inferring effects on associated, unlisted components like runways.

Typical baseline systems, lacking structured knowledge (e.g., airport-runway relationships) or advanced reasoning, often fail this inference. They might parse the airport closure but omit the runway, providing incomplete awareness:
\begin{quote}\small
\texttt{\{"airport": "AGGC", "runway": \textbf{""}, ...\}} % Baseline misses the implied runway link
\end{quote}

Our framework addresses this challenge. KG-TableRAG queries the aviation knowledge graph with the airport identifier ('AGGC'), retrieving that \texttt{"RWY 07R"} belongs to airport \texttt{"AGGC"}. This fact supplies the missing structural context.

The LLM then integrates the input instruction ("airport closed") with this retrieved fact. Through semantic reasoning, it correctly infers the operational consequence - the runway must also be closed because it is part of the closed airport, leading to the accurate output:
\begin{quote}\small
\texttt{\{"airport": "AGGC", "runway": \textbf{"RWY 07R"}, ...\}} % Our model correctly infers runway closure
\end{quote}

This correct inference of runway \texttt{"RWY 07R"} is critical for operational safety (e.g., preventing routing to a closed runway). It highlights our approach's advantage: integrated knowledge and reasoning for comprehensive understanding beyond simple text extraction. For additional detailed examples, please refer to the supplementary material.

\section{Conclusion}

We present NOTAM-Evolve, a self-evolving framework that addresses the fundamental challenges of structured NOTAM interpretation through deep parsing. Unlike existing approaches limited to shallow pattern matching, our framework tackles the dual reasoning requirements of dynamic knowledge grounding and schema-based inference through three synergistic components: KG-TableRAG for knowledge-guided retrieval, iterative self-optimization through preference learning, and multi-view inference with consensus decoding.

Experimental validation on our dataset of 10,000 globally sourced NOTAMs demonstrates significant advancement over existing methods. Our framework achieves a 30.4\% absolute accuracy improvement over the base LLM, establishing new state-of-the-art performance among open-source solutions while approaching commercial model capabilities. This research establishes a new paradigm for automated NOTAM analysis, with principles extensible to other high-precision domains requiring robust knowledge integration and adaptive learning.

\textbf{Limitations and Future Directions.} While demonstrating substantial improvements, our framework faces computational scalability challenges with progressive iteration costs similar to reinforcement learning paradigms. The inherent complexity of NOTAMs makes creating perfectly accurate ground truth annotations challenging, potentially limiting performance ceilings. Future work could explore LLM-assisted annotation combined with expert validation, more efficient optimization strategies, and extensions to multi-lingual NOTAMs and real-time operational scenarios.

\section{Acknowledgments}
This work was supported by the Beijing Natural Science Foundation (JQ24019, QY24215). This work is being supported by the Open Project Program of the State Key Laboratory of CNS/ATM (No. 2024B31). This work was supported in part by the National Natural Science Foundation of China (Grant Nos. 62576047, 52572349).

\bibliography{aaai2026}

\appendix
\section{Training Algorithm Implementation Details}
\label{sec:appendix}

% --- 合并与简化后的算法 ---
\begin{algorithm}[H]
\small
\caption{Unified Self-Evolving Optimization Process}
\label{alg:unified-self-evolve}
\begin{algorithmic}[1] % [1] 开启行号, 每行都编号
\State \textbf{Input:} Initial dataset $\mathcal{D}_0$, base model $\pi_{\text{base}}$, empty response pool $\mathcal{R} = \emptyset$
\State \textbf{Hyperparameters:} Max iterations $T$, error threshold $\tau$, DPO temperature $\beta$, epochs $E$
\State $\pi_{\text{current}} \gets \pi_{\text{base}}$

\For{$t = 1$ to $T$}
    \State \Comment{\textit{Phase 1: Supervised Fine-Tuning (SFT)}}
    \State Generate responses for $\mathcal{D}_0$ using $\pi_{\text{current}}$ and update response pool $\mathcal{R}$
    \State Construct SFT dataset $\mathcal{D}_{\text{SFT}}$ from high-quality responses in $\mathcal{R}$
    \State $\pi_{\text{SFT}} \gets$ Fine-tune $\pi_{\text{current}}$ on $\mathcal{D}_{\text{SFT}}$
    
    \State \Comment{\textit{Phase 2: Direct Preference Optimization (DPO)}}
    \State Generate responses for $\mathcal{D}_0$ using $\pi_{\text{SFT}}$ and update response pool $\mathcal{R}$
    \State Construct preference dataset $\mathcal{D}_{\text{pref}} = \{(x, Y_w, Y_l)\}$ from pairs in $\mathcal{R}$
    
    \If{$\mathcal{D}_{\text{pref}}$ is not empty}
        \State \Comment{\textit{DPO Training with Curriculum Learning}}
        \State Augment high-error samples in $\mathcal{D}_{\text{pref}}$ to create $\mathcal{D}_{\text{aug}}$
        \State Train $\pi_{\text{SFT}}$ for $E$ epochs using the DPO loss ($\mathcal{L}_{\text{DPO}}$) on $\mathcal{D}_{\text{aug}}$
        \State \quad \textit{(Apply adaptive weights to prioritize high-error samples during training)}
        \State $\pi_{\text{current}} \gets \text{resulting model } \pi_{\text{DPO}}$
    \Else
        \State $\pi_{\text{current}} \gets \pi_{\text{SFT}}$ \Comment{Skip DPO if no new preference data is found}
    \EndIf
\EndFor
\end{algorithmic}
\end{algorithm}

\section{Task Prompt}

\begin{promptbox}{NOTAM Runway Status Parsing Prompt}
As an AI assistant specialized in parsing NOTAMs, extract \textbf{runway status information} according to the following structured rules.

\textbf{Scope:} Focus only on runway closure, restriction, or reopening messages.  
Ignore taxiway, apron, or lighting-related NOTAMs.

\vspace{1ex}
\textbf{Runway Status Classification:}
\begin{itemize}[leftmargin=*]
    \item \textbf{Closed (MRLC, MRXX):} Keywords include \texttt{CLOSED}, \texttt{CLSD}, \texttt{CLOSURE}, \texttt{NOT AVBL}, \texttt{UNAVAILABLE}, \texttt{SUSPENDED}, etc.
    \item \textbf{Limited / Restricted (MRLT, MRXX):} Keywords include \texttt{RESTRICTED}, \texttt{LIMITED}, \texttt{RESERVED FOR}, often combined with “only”.
    \item \textbf{Open / Cancellation (MRAH):} Keywords include \texttt{OPEN}, \texttt{OPN TO TFC}, \texttt{CANCELLED CLOSURE}, etc.
\end{itemize}

\vspace{1ex}
\textbf{Impact Evaluation:}
\begin{itemize}[leftmargin=*]
    \item Determine whether the restriction affects takeoffs, landings, or both.  
    If unspecified, assume both.
    \item Identify affected flight types:
    \begin{itemize}[leftmargin=*]
        \item If not mentioned: assign “International, Domestic, Regional”.
        \item If explicitly mentioned (e.g., “INTERNATIONAL FLIGHT ONLY”): assign accordingly.
    \end{itemize}
    \item Identify affected aircraft types:
    \begin{itemize}[leftmargin=*]
        \item If wingspan, CODE (e.g., C/D), or engine number is mentioned, fill in \texttt{affect\_actype}.
        \item Convert wingspan from FT to M if required.
    \end{itemize}
\end{itemize}

\vspace{1ex}
\textbf{Output Format:}
\begin{itemize}[leftmargin=*]
    \item \texttt{airport}: ICAO code
    \item \texttt{runway}: Runway number
    \item \texttt{affect\_actype}: Affected aircraft type or \texttt{null}
    \item \texttt{affect\_region}: TAKEOFFS | LANDINGS | TAKEOFFS,LANDINGS
    \item \texttt{flight\_type}: International | Domestic | Regional (use Chinese wording if applicable)
\end{itemize}

\vspace{1ex}
\textbf{Notes:}
\begin{itemize}[leftmargin=*]
    \item Partial closure/restriction = full closure/restriction.
    \item Create separate records for each runway mentioned.
    \item Extract only explicitly stated information; avoid assumptions.
    \item Preserve Chinese wording for flight types when present.
    \item Match CODE or category restrictions with aircraft type table if needed.
\end{itemize}

\vspace{1ex}
Now, based on the above rules, extract relevant information from the given NOTAM text and output in JSON format.
\end{promptbox}

\section{Example}
\label{sec:example}

\begin{promptbox}{Example Input}
\texttt{
<instruction>:
Q)KZDV/QMRLC/IV/NBO/A/000/999/
3952N10440W005
A)KDEN B)2301010254 C)2301011200
E) DEN RWY 17L/35R CLSD
}
\end{promptbox}

\begin{promptbox}{Example Model Output}
\texttt{<result:>\newline
<think>\newline
Runway Identification:\newline
    Extracted ICAO code "KDEN" from "A) KDEN".\newline
\newline
Runway Status Analysis:\newline
    "RWY 17L/35R CLSD" → both runways fully closed.\newline
    Default impact: both takeoffs and landings → affect\_region = "TAKEOFFS,LANDINGS".\newline
\newline
Aircraft Type Restriction:\newline
    None mentioned → affect\_actype = null.\newline
\newline
Flight Type Inference:\newline
    No specific restriction → flight\_type = "International,Domestic,Regional".\newline
\newline
Multiple Runways Handling:\newline
    Separate records for 17L and 35R.\newline
</think>\newline
\newline
[\newline
  \{ "airport": "KDEN", "runway": "17L", "affect\_actype": null, "affect\_region": "TAKEOFFS,LANDINGS", "flight\_type": "International,Domestic,Regional" \},\newline
  \{ "airport": "KDEN", "runway": "35R", "affect\_actype": null, "affect\_region": "TAKEOFFS,LANDINGS", "flight\_type": "International,Domestic,Regional" \}\newline
]}%
\end{promptbox}

\end{document}